\pdfoutput=1
\documentclass[11pt]{article}

\usepackage[final]{acl}

\usepackage{times}
\usepackage{latexsym}

\usepackage[T1]{fontenc}
\usepackage[utf8]{inputenc}

\usepackage{microtype}

\usepackage{inconsolata}

\usepackage{amsmath}
\usepackage[ruled, boxed, linesnumbered]{algorithm2e}
\usepackage[pdftex]{graphicx}
\usepackage{subcaption}
\usepackage{wrapfig}
\usepackage{multirow}
\usepackage{booktabs}
\usepackage{xcolor}
\usepackage{caption}
\usepackage{url}

\title{CQIL: Inference Latency Optimization with Concurrent \protect\\ Computation of Quasi-Independent Layers}

\author{
    Longwei Zou\textsuperscript{\normalfont 1}, 
    Qingyang Wang\textsuperscript{\normalfont 2},
    Han Zhao\textsuperscript{\normalfont 3}, \\
    {\bf Jiangang Kong\textsuperscript{\normalfont 3}}, 
    {\bf Yi Yang\textsuperscript{\normalfont 3}}, 
    {\bf Yangdong Deng\textsuperscript{\normalfont 1}} \\
    \textsuperscript{1}Tsinghua University, \textsuperscript{3}DiDi Global Inc, \\
    \textsuperscript{2}BNU-HKBU United International College \\
    \texttt{zoulw22@mails.tsinghua.edu.cn}, \texttt{q030026149@mail.uic.edu.cn} \\ 
    \texttt{\{zhaohan,kongjiangang,yangyiian\}@didiglobal.com} \\
    \texttt{dengyd@tsinghua.edu.cn}
}

\begin{document}
\maketitle
\begin{abstract}
The fast-growing large scale language models are delivering unprecedented performance on almost all natural language processing tasks. However, the effectiveness of large language models are reliant on an exponentially increasing number of parameters. The overwhelming computation complexity incurs a high inference latency that negatively affects user experience. Existing methods to improve inference efficiency, such as tensor parallelism and quantization, target to reduce per-layer computing latency, yet overlook the cumulative latency due to the number of layers. Recent works on reducing the cumulative latency through layer removing, however, lead to significant performance drop. Motivated by the similarity of inputs among adjacent layers, we propose to identify quasi-independent layers, which can be concurrently computed to significantly decrease inference latency. We also introduce a bypassing technique to mitigate the effect of information loss. Empirical experiments of the proposed approach on the LLaMA models confirm that Concurrent Computation of Quasi-Independent Layers (CQIL) can reduce latency by up to 48.3\% on LLaMA-33B, while maintaining a close level of performance. \footnote{Code is available at https://github.com/Photooon/CQIL}

\end{abstract}

\section{Introduction}

Large Language Models (LLMs) are offering unprecedented power to deliver remarkable performance across diverse tasks of natural language processing. The exceptional performance, however, comes at the cost of increasing model size and, consequently, higher inference latency. For example, the per-token inference time for GPT-4 is approximately three times longer than that of GPT-3.5, according to measurements from the OpenAI API \footnote{Note that latency can vary depending on time and location; the reported times is based on observations at the authors' location.}. Such a high latency has a direct impact on user experience, highlighting the urgent need to mitigate inference delays of LLMs.

\begin{figure}[t]
    \centering    
    \includegraphics[width=\columnwidth]{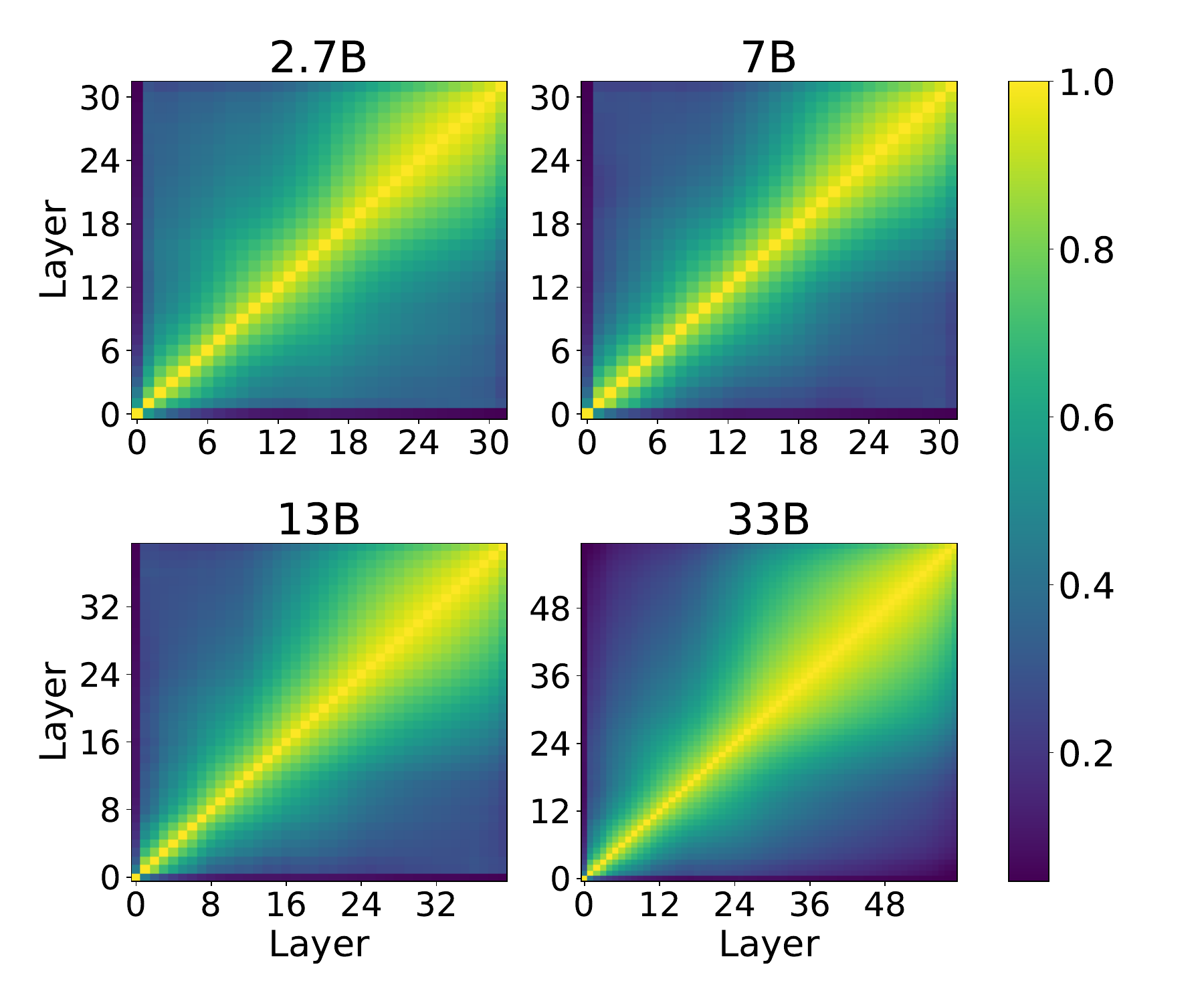}
    \caption{Similarity of inputs across layers in LLaMA-1 models. Sub-figure with title "2.7B" represents the similarity of inputs in Sheared-LLaMA-2.7B\cite{sheared-llama}. It highlights that adjacent layers have highly similar input. Notably, such similarity of inputs becomes increasingly evident in larger models at deeper layers, suggesting the quasi-independence of deeper layers and the potential for parallel computation.}
    \label{fig:layer_input_sim}
\end{figure}

LLMs typically consist of a large number of sequentially connected layers with identical structures. For better illustration, we analyze the inference latency of LLMs along two dimensions: per-layer latency and the cumulative latency due to all layers. Existing low-latency inference methods, like tensor parallelism, quantization \cite{LLM.int8(), SmoothQuant}, and unstructured pruning \cite{DeepCompression, SparseGPT, Wanda} and low-rank factorization \cite{ALBERT, LightFormer}, primarily focus on minimizing per-layer latency. Meanwhile, there are works, such as structured pruning \cite{LLM-Pruner, sheared-llama} and dynamic early existing \cite{DeeBERT}, proposed to address the cumulative latency through removing and/or dynamically omitting layers. These approaches, unfortunately, often result in considerable performance degradation and potential loss of learned knowledge \cite{Key-Value-Memories}. Therefore, it's critical to develop effective methods that can reduce the latency related to the total number of layers while preserving the model performance.

This work is inspired by the observation that adjacent layers in LLMs share significantly similar inputs. Such similarity suggests the possibility of substituting a layer's input with that of a certain preceding layer without significantly altering its output. Such layers sharing input are designated as quasi-independent layers in this work. As a result, the computing dependency between adjacent layers can be eliminated and thus unleash the potential of parallel computation. We introduce a framework of Concurrent Computation of Quasi-Independent Layers (CQIL), to reduce LLM inference latency by parallelizing the computation across layers with similar inputs. Additionally, we develop a bypassing technique to transmit the output of attention modules among input-aligned layers, with the purpose of minimizing the information loss. Extensive experiments demonstrate reductions of inference latency by up to 48.3\% on the LLaMA-33B model, with a minor impact on performance. We also discuss the implication of CQIL in the context of ensembles, which may offer deeper insights into the fundamental characteristic of LLMs.

The major contributions of this work are as follows. First, we propose CQIL, a novel approach to enhance the inference efficiency of LLMs through concurrent computation of quasi-independent layers, effectively addressing the challenge posed by the increasing number of layers. Second, our method enables the adaptation of pre-trained LLMs into ensemble-like models with minimal performance loss, which may provide deeper insights for layers' characters in LLMs. Third, We effectively reduce the inference latency of LLaMA models, with minimal impact on the model performance.

\section{Related Work}

\paragraph{Efficient Inference Approaches} Model compression techniques, such as pruning \cite{SparseGPT, Wanda, LLM-Pruner, sheared-llama}, quantization \cite{LLM.int8(), SmoothQuant}, low-rank factorization \cite{ALBERT, LightFormer}, and knowledge distillation \cite{KD, DistilBERT, TinyBERT, MobileBERT}, reduce inference latency by trimming parameters in the model. Methods like dynamic early exit \cite{DeeBERT} and speculative decoding \cite{SpeculativeDecoding, SpeculativeSampling} leverage intermediate layer output and output from smaller models to predict the final outcome ahead of time. Flash Attention \cite{flash_attention} enhances the efficiency of attention computation by carefully orchestrating computation and memory usage. These methods are orthogonal to our approach, allowing for potential integration for a higher level of improvement. Notably, we test the combination of the pruning approach with our method, detailed in Section \ref{sec:combined_with_pruning}.

\paragraph{Parallelism} Beyond removing the model's parameters, latency reduction and throughput enhancement can also be achieved through parallel computation strategies. Typically, parallelism in LLM computation includes data parallelism \cite{DeepSpeed}, pipeline parallelism \cite{PipeDream, GPipe}, and tensor parallelism. Among these, only tensor parallelism addresses the inference latency by distributing layers computation across multiple GPUs. Similar to tensor parallelism, our method employs additional GPUs to decrease latency. Therefore, we conducted comparison experiments in Section \ref{sec:compared_with_tp}. Results show that, compared with tensor parallelism, our method consistently reduces latency across various batch size, making it more amenable to the scenario of online inference. In addition, our approach is orthogonal to tensor parallelism and we could achieve further acceleration by integrating tensor parallelism with CQIL.

\begin{figure*}[t]
    \centering
    \includegraphics[width=\textwidth]{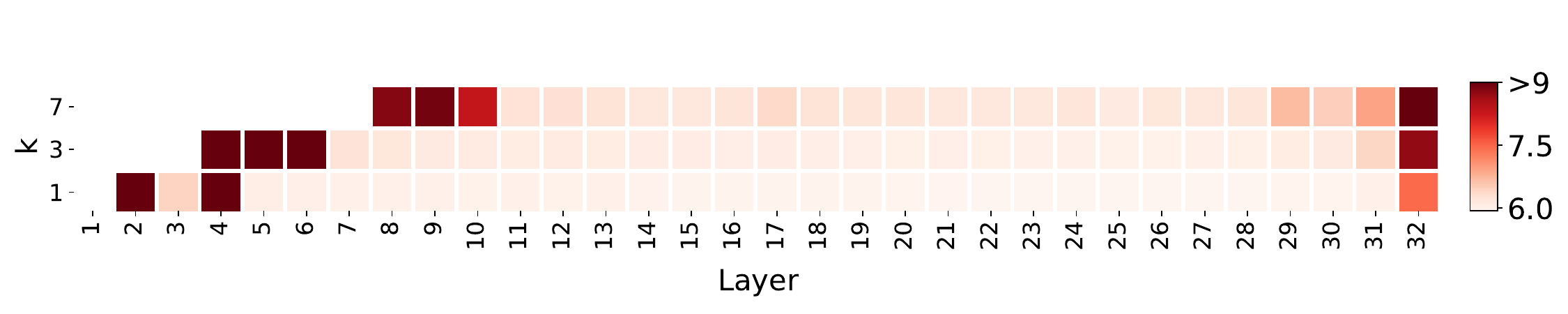}
    \caption{Sensitivity of Input Substitution. We individually replace the input of layer $l$ with that of the layer $l-k$ and evaluate the perplexity. A darker block indicates a higher perplexity and diminished performance. When $k \ge l$, there is no corresponding layer for $l-k$, therefore these parts are left blank in the figure. The original perplexity is around 6. The drawing shows that both bottom and top layers (bottom refers to the direction close to the embedding layer) are sensitive to the input substitution, whereas the majority of middle layers are relatively insensitive.}
    \label{fig:misin}
\end{figure*}

\paragraph{Transformers with Parallel Architecture} Previous works have explored the acceleration of pre-training phases through the parallelization of transformer architectures. GPT-J \cite{GPT-J} and PaLM \cite{PaLM} achieve efficiency improvements by parallelizing the attention and feedforward modules within transformer layers, allowing concurrent computation for the beginning projection in these two modules. Other research efforts \cite{multi-pass, cross-paralle} have enhanced model performance by expanding model width through parallel layers design, and still focus on the pre-training stage. Our work applies concurrent computation of layers to a pre-trained LLMs, aiming at reducing latency while maintaining model performance. Our approach thus differs from existing methods that focus on pre-training transformers with parallel architecture in terms of both motivation and methodology.

\section{Preliminary}

In this section, we present a systematic investigation that motivates the design of the proposed approach introduced in Section \ref{sec:methodology}. Our discussion proceeds under the assumption of pre-layer normalization, which is adopted by most LLMs \cite{GPT3, OPT, LLaMA}. In following experiments, we utilize the LLaMA-1 \cite{LLaMA} and Sheared-LLaMA-2.7B \cite{sheared-llama} models, with the input samples randomly selected from the RedPajama \cite{redpajama} dataset. We first investigate the similarity of inputs across layers in LLMs, which suggests the quasi-independence of deeper layers and the potential for parallel computation. Second, we explore the effect of substituting a layer's input with that of its preceding layers, which aids in identifying potential layers for parallel processing with minor performance degradation.

\subsection{Similarity of Layer Input}

For a LLM with $L$ layers, we define the input to the layer $l$ as $x_l \in \mathbf{R}^{B, T, H}$, where $B$, $T$, and $H$ denote batch size, token count, and hidden dimension size, respectively. The output from the layer $l$ is denoted as $x_{l+1} = x_{l} + F_l(x_{l})$. We employ cosine similarity to quantify the similarity of inputs between layers.

Figure \ref{fig:layer_input_sim} reveals the similarity of inputs across layers in LLaMA models with different parameter sizes. The following two primary observations emerge from our experiment results. First, input similarity intensifies with increasing depth within the model, attributed to the cumulative effect of pre-layer normalization, which makes the input difference $F_{l-1}(x_{l-1})$ between adjacent layers $l-1$ and $l$ increasingly negligible compared with the cumulative value $x_l = x_1 + F_1(x_1) + ... + F_{l-1}(x_{l-1})$. Second, larger models exhibit more obvious similarity of inputs, suggesting that layers in such models are quasi-independent and parallel computation may be more readily available in such models.

While prior research \cite{CRaSh, LRJump} has identified output similarity across layers to facilitate inference efficiency through pruning, our approach is based on a different motivation with a more systematic perspective. Experimental results of previous works indicate that layer pruning often result in substantial performance declines. For instance, although the Sheared-LLaMA-1.3B model has a similar hidden dimension size to that of the Sheared-LLaMA-2.7B, it contains eight fewer layers and consequently results in significantly performance drop on downstream tasks. Contrary to removing layers, our approach seeks to parallelize layer computation to reduce inference latency while preserving the model performance. In addition, our method is orthogonal to pruning techniques and the compatibility is further demonstrated in Section \ref{sec:combined_with_pruning}.

We then empirically assess the effect of substituting each layer's input with that of preceding layers, facilitating the identification of layers suitable for parallel computation.

\subsection{Sensitivity of Input Substitution}
\label{sec:substituted_input_tolerance}

\begin{figure*}[t]
    % \vspace{0.25cm}
    % \setlength{\belowcaptionskip}{0.25cm}
    \centering    
    \includegraphics[width=0.85\textwidth]{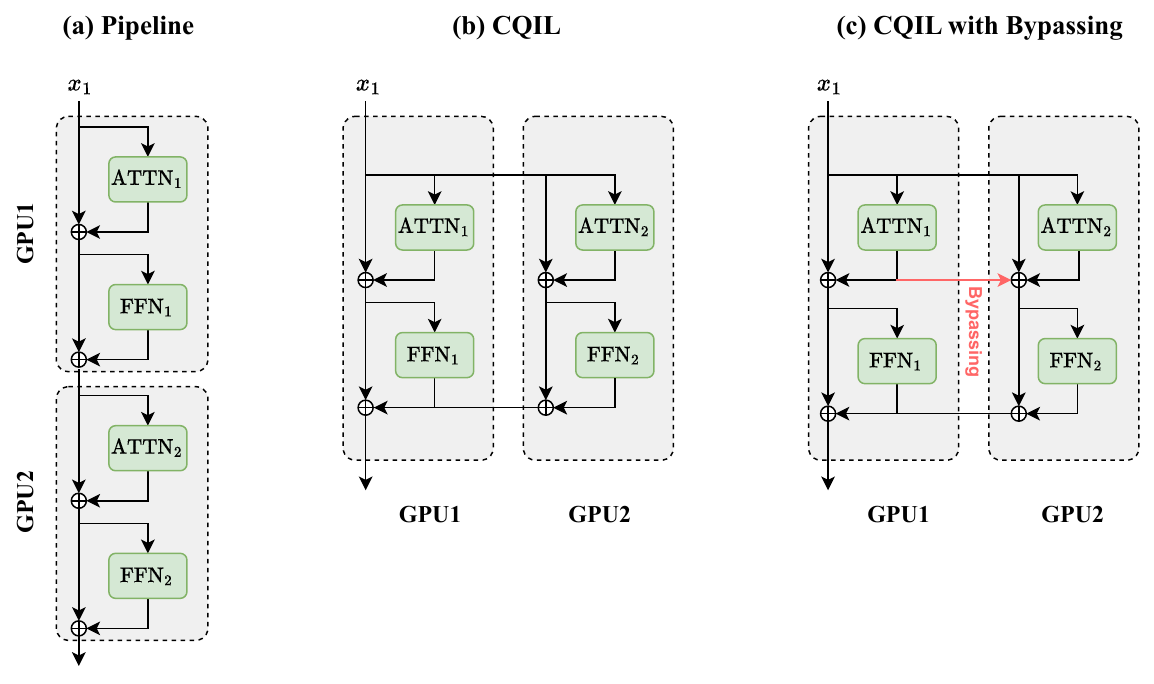}
    \caption{The proposed method. (a-c) depict the pipeline inference as well as the CQIL with and without the bypassing technique. The pipeline inference represents the standard setup, where layers are processed sequentially. In contrast, CQIL substitutes the input for layer $2$ with that of layer $1$, enabling concurrent computation across two GPUs for latency reduction. Given that both layers produce attention outputs concurrently, and the attention output of layer $1$ serves as an input for layer $2$ in the original model, the bypassing technique transmits the attention output from GPU1 to GPU2, minimizes the information loss and improves the model performance.}
    \label{fig:method}
\end{figure*}

As shown in Figure \ref{fig:layer_input_sim}, adjacent layers have similar inputs. It is thus appealing to identify possible layers for parallel computation, i.e., quasi-independent layers. In this section, we substitute the input of a layer with that of preceding layers to evaluate the sensitivity of each layer and thus justify the feasibility of quasi-independent layers. Specifically, the output of the model can be written as $x_{L+1} = x_1 + F_1(x_1) + ... F_l(x_{l}) + ... F_L(x_{L})$. For layer $l$, we replace the input of $F_l$ with $x_{l-k}$, which is the input of layer $l-k$, and keep other terms unchanged. This process is repeated for every layer individually, assessing their adaptability to input changes through perplexity measurements on validation set. When perplexities of the original and the input-substituted models are similar, it is feasible to concurrently compute these two layers. 

As illustrated in Figure \ref{fig:misin}, our substitution trials with the LLaMA-7B model indicate that the input for a majority of the middle layers can be effectively replaced by those of their immediate predecessors while maintaining a similar performance. It is observed that both the bottom and top layers are sensitive to input changes. The bottom layers display less input similarity compared to subsequent layers. In addition, substitution in bottom layers leads to a propagation of errors from substituted inputs through subsequent layers, thereby amplifying their susceptibility to input changes. The top layers, due to their direct connection with the output distribution, also experience a notable impact on performance if their input are substituted.

The empirical findings suggest that most middle layers offer a tolerance for input substitution. In other words, it is feasible to relax the computing dependence by using the same input to quasi-independent layers, revealing the potential of parallel computation for latency reduction.

\section{Methodology}
\label{sec:methodology}

\subsection{Problem Statement}

Given a LLM with pre-layer normalization, we can formulate the computation of layer $l$ as Eq. \ref{eq:input_formulation} and Eq. \ref{eq:attn_and_ffn}. ATTN and FFN represent the attention and feedforward modules in each layer. $n$ denotes the number of heads, and $d_k$ specifies the dimension size of each head.

\begin{align}
    x_{l+1} = x_{l} &+ \text{ATTN}_{l}(x_{l}) + \text{FFN}_{l}(x_{l} + \text{ATTN}_{l}(x_{l}))
    \label{eq:input_formulation}
\end{align}

\begin{align}
    \text{ATTN}(x) &= \text{MHA}(\text{Norm}(x)) \notag \\
    \text{MHA}(x) &= \text{Concat}(\text{head}_{\text{1}}, ..., \text{head}_{\text{n}})\boldsymbol{W}^O \notag \\
    \text{head}_{\text{i}} &= \text{Attention}(x\boldsymbol{W}_i^Q, x\boldsymbol{W}_i^K, x\boldsymbol{W}_i^V) \notag \\
    \text{Attention}&(Q, K, V) = \text{softmax}(\frac{Q K^T}{\sqrt{d_k}})V \notag \\
    \text{FFN}(x) &= \boldsymbol{W}_2 f(\boldsymbol{W}_1 (\text{Norm}(x)) + b_1) + b_2 \notag \\
    \label{eq:attn_and_ffn}
\end{align}

Our objective is to replace the input $x_{l}$ in Eq. \ref{eq:input_formulation} with $x_{i}, i<l$. This substitution implicitly enables the parallel computation of layers $l$ and $i$ and thus reduces the inference latency.

\subsection{Layer Partition and Parallelism}

% \IncMargin{-0.1em}
\begin{algorithm}[t]
    \caption{Concurrent Computation of Quasi-Independent Layers}
    \label{alg:cqil}
    \SetKwInOut{Input}{Input}
    \SetKwInOut{Output}{Output}
    \SetKwRepeat{for}{do}{end}
    
    \Input{input $x_1$, model with $L$ layers, group size $p$, start layer $s$, end layer $e$ and bypassing distance $d$.}

    \Output{output $x_L$}

    \For{$l = 1 \rightarrow s$, step=$1$}{
        $a_{l} = x_{l} + \text{ATTN}_{l}(x_{l})$ \\
        $x_{l+1} = a_{l} + \text{FFN}_{l}(a_{l})$ \\
    }

    \For{$l = s+1 \rightarrow e$, step=$p$}{
        reqs1, reqs2, attns = [], [], [] \\
        \For{$i = 0 \rightarrow p-1$, step=$1$}{
            reqs1.add(non\_block\_exec( \\
                $a_{l+i, attn} = \text{ATTN}_{l+i}(x_{l})$, gpu=$i$)) \\
        }
        
        \For{$i = 0 \rightarrow p-1$, step=$1$}{
            reqs1[i].wait() \\
            attns.add($a_{l+i, attn}$) \\
            $bp = \text{sum}(\text{attns}[-\text{min}(d,i)+1:])$ \\
            reqs2.add(non\_block\_exec( \\
                $a_{l+i, ffn} = \text{FFN}_{l+i}(x_{l} + bp)$, gpu=$i$)) \\
        }
        $x_{l+p} = x_{l}$ \\
        \For{$i = 0 \rightarrow p-1$, step=$1$}{
            reqs2[i].wait() \\
            $x_{l+p} = x_{l+p} + a_{l+i, attn} + a_{l+i, ffn}$ \\
        }
    }
    % // last layers are computed as beginning \\
    \For{$l = e + 1 \rightarrow L$, step=$1$}{
        $a_{l} = x_{l} + \text{ATTN}_{l}(x_{l})$ \\
        $x_{l+1} = a_{l} + \text{FFN}_{l}(a_{l})$ \\
    }
\end{algorithm}
% \DecMargin{-1em}
% \vspace{-1em}

% TODO: symbol那个as

Considering a LLM with $L$ layers, we partition layers into $K$ groups $G_1, G_2, ..., G_K$. Within each group, layers share the same input, allowing for concurrent computation of them. The computation for group $G_{k}$, where $1 \leq k \leq K$, can be formalized as shown in Eq. \ref{eq:grouped_input_formulation}. ATTN and FFN operations within each summation in this equation are executed in parallel. Specifically, when the size of each group is $1$, Eq. \ref{eq:grouped_input_formulation} reverts to the sequential computation in Eq. \ref{eq:input_formulation}.

\begin{align}
    x_{G_{k+1}} = x_{G_{k}} + & \sum_{l \in G_{k}}{\text{ATTN}_{l}(x_{G_{k}})}  \notag \\
                    + & \sum_{l \in G_{k}}{\text{FFN}_{l}(x_{G_{k}} + \text{ATTN}_{l}(x_{G_{k}}))}
    \label{eq:grouped_input_formulation}
\end{align}

The number of potential partition schemes can be considerable. Therefore, we employ a straightforward approach to divide layers into groups. Specifically, with a maximum group size of $p$, start layer $s$, and end layer $e$, we sequentially arrange layers from $s$ to $e$ into groups of size $p$, while organizing remaining layers into groups of size $1$. For instance, the LLaMA-7B model with $32$ layers, can be segmented into groups as $\{1\} \to \{2\} \to ... \to \{8\} \to \{9,10\} \to \{11,12\} \to ... \to \{29,30\} \to \{31\} \to \{32\}$, with $p=2, s=9, e=30$. Such partitioning strategy helps prevent low GPU utilization and significant performance degradation from parallelizing the bottom and top layers. Algorithm \ref{alg:cqil} further expound the concurrent computation process. Hyperparameter $p$ is determined according to the number of GPUs. The choice of hyperparameters $s, e$ depends on balancing performance against acceleration, discussed further in Section \ref{sec:trade_off}.

\subsection{Bypassing}

\begin{table*}[t]
    \centering
    \caption{Downstream tasks performance and latency reduction. $p=1$ refers to the original model. Results indicate that our method effectively reduce the inference latency of LLaMA models, while preserving the model performance. Additionally, as the model size and the number of GPUs increases, CQIL achieves further latency reductions.}
    \resizebox{\textwidth}{!}{
        \begin{tabular}{cccccccccc}
            \toprule
            \multirow{2}{*}{\begin{tabular}[c]{@{}c@{}} \textbf{Model} \end{tabular}} & \multicolumn{3}{c}{\textbf{Partition Strategy}} & \multicolumn{6}{c}{\textbf{Commonsense\&Reading Comprehension}} \\
            % \cline{3-8}
             & $p$ & $s$ & $e$ & \textbf{SciQ} & \textbf{PIQA} & \textbf{WinoGrande} & \textbf{ARC-E} & \textbf{ARC-C} & \textbf{HellaSwag} \\
            \hline
            \multirow{3}{*}{\begin{tabular}[c]{@{}c@{}} \textbf{LLaMA-7B} \end{tabular}} 
             & 1 & 1 & 32 & 93.0 & 79.2 & 70.0 & 72.9 & 44.8 & 76.2 \\
             & 2 & 13 & 30 & 90.8 & 78.3 & 69.0 & 70.1 & 41.3 & 73.5 \\
             & 4 & 15 & 30 & 89.5 & 76.0 & 68.3 & 65.5 & 39.6 & 71.0 \\
            \hline
            \multirow{3}{*}{\begin{tabular}[c]{@{}c@{}} \textbf{LLaMA-13B} \end{tabular}} 
             & 1 & 1 & 40 & 91.3 & 80.1 & 72.8 & 74.8 & 47.6 & 79.1 \\
             & 2 & 11 & 38 & 89.3 & 79.4 & 69.6 & 72.1 & 44.9 & 76.8 \\
             & 4 & 15 & 38 & 90.7 & 78.8 & 70.2 & 71.0 & 43.7 & 75.5 \\
             \hline
            \multirow{3}{*}{\begin{tabular}[c]{@{}c@{}} \textbf{LLaMA-33B} \end{tabular}} 
             & 1 & 1 & 60 & 94.6 & 82.3 & 76.0 & 79.0 & 52.1 & 82.6 \\
             & 2 & 11 & 58 & 94.5 & 80.4 & 71.8 & 76.1 & 51.5 & 80.9 \\
             & 4 & 19 & 58 & 94.0 & 79.4 & 74.7 & 75.0 & 50.1 & 80.8 \\
             \midrule
            \multirow{2}{*}{\begin{tabular}[c]{@{}c@{}} \textbf{Model} \end{tabular}} & \multicolumn{3}{c}{\textbf{Partition Strategy}} & \multicolumn{2}{c}{\textbf{Continued}} & \textbf{LM} & \textbf{World Knowledge} & \textbf{Downstream Tasks} & \textbf{Latency} \\
              & $p$ & $s$ & $e$ & \textbf{LogiQA} & \textbf{BoolQ} & \textbf{LAMBADA} & \textbf{MMLU (5)} & \textbf{Average Score} & \textbf{Reduction} \\
            \hline
            \multirow{3}{*}{\begin{tabular}[c]{@{}c@{}} \textbf{LLaMA-7B} \end{tabular}}
             & 1 & 1 & 32 & 30.0 & 75.1 & 73.5 & 35.1 & 65.0 & 0\% \\
             & 2 & 13 & 30 & 29.0 & 74.4 & 72.9 & 32.9 & 63.2 & 27.0\% \\
             & 4 & 15 & 30 & 29.5 & 72.6 & 69.9 & 33.2 & 61.5 & 36.0\% \\
            \hline
            \multirow{3}{*}{\begin{tabular}[c]{@{}c@{}} \textbf{LLaMA-13B} \end{tabular}} 
             & 1 & 1 & 40 & 32.0 & 77.9 & 76.2 & 46.7 & 67.8 & 0\% \\
             & 2 & 11 & 38 & 29.3 & 74.6 & 73.7 & 40.7 & 65.0 & 34.0\% \\
             & 4 & 15 & 38 & 30.3 & 76.8 & 73.7 & 43.1 & 65.4 & 43.2\% \\
             \hline
            \multirow{3}{*}{\begin{tabular}[c]{@{}c@{}} \textbf{LLaMA-33B} \end{tabular}} 
             & 1 & 1 & 60 & 31.8 & 82.6 & 77.6 & 58.2 & 71.7 & 0\% \\
             & 2 & 11 & 58 & 28.9 & 80.5 & 76.9 & 50.5 & 69.2 & 38.6\% \\
             & 4 & 19 & 58 & 30.1 & 81.0 & 75.4 & 52.0 & 69.3 & 48.3\% \\
            \bottomrule
        \end{tabular}
    }
    \label{tab:experiment_results}
    \vspace{-0.25cm}
\end{table*}

The transformer layer includes an attention module and a feedforward module. Notably, parallel computation of $p$ layers in group $G_k$ allows concurrent accesses to attention module outputs $\text{ATTN}_{l}(x_{G_k}), l \in G_k$. We could transfer $\text{ATTN}_{l}(x_{G_k})$ to the feedforward module of layer $j$, $j > l$ and $j \in G_k$, thereby minimizing information loss. We designate such a approach as bypassing by following the terminology of computer architecture. We define the bypassing distance $d$ as the maximum of $j - l$. With bypassing, the formulation for $G_{k+1}$ becomes Eq. \ref{eq:bypassing_input_formulation}. 

\begin{align}
    \begin{split}
    x_{G_{k+1}} = x_{G_{k}} &+ \sum_{l \in G_{k}}{\text{ATTN}_{l}(x_{G_{k}})} \\
    &+ \sum_{l \in G_{k}}{\text{FFN}_{l}( 
    x_{G_{k}} + \text{ATTN}_{l}(x_{G_{k}})} \\
    &+ {\sum_{l' \in G_{k}, 1 \leq l-l' \leq d}{\text{ATTN}_{l'}(x_{G_{k}})})}
    \label{eq:bypassing_input_formulation}
    \end{split}
\end{align}

Moreover, the number of transmissions for bypassing is ${d(2p -d - 1)}/{2}$. To avoid communication bottlenecks, we set $d=1$ across all experiments unless otherwise specified. The effect of bypassing distances are detailed in Section \ref{sec:bypassing_expr}.

% 给表格最后两列的数字加粗

\subsection{Fine-tuning}

Without additional training, models with concurrent computation of quasi-independent layers maintain a close level of performance to the original. To achieve better performance, we fine-tune the model using LoRA\cite{lora} on the pre-training dataset. Fine-tuning with just 0.5B tokens is sufficient to almost reach the original performance, suggesting that middle layers in original LLMs may inherently function as ensembles. We discuss such implication further in Section \ref{sec:discussion}. We believe that continual pre-training with more tokens will yield further performance enhancements. However, due to computational resource constraints, we leave the continual pre-training for the model with CQIL as future work.

% TODO: 需要添加一个附录，给没finetune的结果，在这里添加链接

\section{Experiment}

\subsection{Experimental Setup}

\paragraph{Model configurations and Dataset} We apply CQIL on Sheared-LLaMA-1.3B, Sheared-LLaMA-2.7B \cite{sheared-llama}, LLaMA-7B, LLaMA-13B and LLaMA-33B \cite{LLaMA}. Concurrent computation of layers are implemented with Pytorch distributed communication package \cite{Pytorch} and Huggingface Transformers library \cite{Huggingface}. We use RedPajama \cite{redpajama}, the replicated dataset of LLaMA1 models, as the training dataset, and adopt the same data mixture used for training LLaMA1. We construct a held-out validation dataset with 500 sequences of 2,048 tokens. 

\paragraph{Baselines} Our method is orthogonal to most existing works on efficient inference, making it hard to be directly compared with them. Given that both tensor parallelism and our strategy employ additional GPUs to reduce latency, we evaluate the relative efficiency of our method against tensor parallelism. The effectiveness of tensor parallelism implementations can vary significantly. Based on empirical assessments, we select the DeepSpeed inference library \cite{DeepSpeed} as the benchmark for tensor parallelism. DeepSpeed inference library partitions attention and feedforward projection matrices and computes them distributively, which has proven to reduce latency in our testing environment. Furthermore, we explore combination of our method and the pruning technique, showing further potential for latency reduction.

\paragraph{Training} We fine-tune models using LoRA on Nvidia A100 GPUs (80GB). Details of fine-tuning are demonstrated in Appendix \ref{sec:finetune_details}. 

\paragraph{Evaluation} We evaluate the downstream tasks performance with lm-evaluation-harness package \cite{eval-harness}. We evaluate 0-shot accuracy of SciQ \cite{SciQ}, PIQA \cite{PIQA}, Winogrande \cite{WinoGrande}, ARC Easy, ARC Challenge \cite{ARC}, HellaSwag \cite{HellaSwag}, LogiQA \cite{LogiQA}, BoolQ \cite{BoolQ}, and LAMBADA \cite{LAMBADA}. We show accuracy of 5-shot MMLU \cite{MMLU}. For the latency, we measure the inference speed on a machine with 8 Nvidia A100 GPUs connected by NVLink and report the average latency reduction across batch sizes ranging from 1 to 256.

% \paragraph{Instruction tuning evaluation} To evaluate ...

\subsection{Downstream Tasks Performance and Latency Reduction}

Table \ref{tab:experiment_results} shows that our method maintains a close level of performance to that of the original model while achieving significant latency reduction. LLaMA-13B and LLaMA-33B utilizing $p=4$ outperform those with $p=2$, attributed to employing a larger $s$ which keeps more bottom layers unchanged.  With $p=4$, our method processes four layers in parallel, achieving further latency improvements over $p=2$. Additionally, as the model size increases, our method demonstrates further latency reductions. This improvement is attributed to the increased quasi-independence of layers observed in larger models.

% \subsection{Instruction Tuning Results}

\subsection{Effect of Bypassing}
\label{sec:bypassing_expr}

\begin{table}[t]
    \centering
    \caption{Effect of bypassing. The performance on downstream tasks gradually improves as $d$ increases. Due to the communication cost, there is a slight decline in latency reduction as $d$ grows.}
    \resizebox{\columnwidth}{!}{
        \begin{tabular}{c|c|c}
            \toprule
            \textbf{Bypassing} & \textbf{Downstream Tasks} & \textbf{Latency} \\
            \textbf{Distance ($d$)} & \textbf{Average Score} & \textbf{Reduction} \\
            \hline
             0 & 60.9 & 35.9\% \\
             1 & 61.2 & 35.7\% \\
             2 & 61.4 & 35.6\% \\
             3 & 61.5 & 34.9\% \\
            \bottomrule
        \end{tabular}
    }
    \label{tab:effect_of_bypassing}
    % \vspace{-0.4cm}
\end{table}

Table \ref{tab:effect_of_bypassing} elucidates the effect of bypassing distances on LLaMA-7B with $p=4,s=14,e=29$. The results suggest that increasing bypassing distance $d$ leads to a gradual improvement of performance, attributed to diminished information loss. Meanwhile, there's a slight decrease in latency reduction as the communication cost rises with larger $d$. However, even at $d=3$, the additional communication cost associated with the bypassing method is still minor in our testing environment. We also recognize that the communication cost depends on the speed of links between GPUs. Therefore, we suggest to select $d$ based on the inference environment. 

\subsection{Trade-Off between Performance and Acceleration}
\label{sec:trade_off}

CQIL involves three hyperparameters, $p$, $s$, and $e$. The selection of $p$ is based on the number of available GPUs. Choices for $s$ and $e$ depend on balancing performance against acceleration. As shown in Table \ref{tab:trade_off}, the performance on downstream tasks and latency reduction vary consistently along with $s$ and $e$. A larger $s$ and a smaller $e$, thereby more layers are unchanged, lead to improved performance and decrease in latency. We suggest to select $s$ and $e$ according to the balance between performance and latency reduction.

\begin{table}[t]
    \centering
    \caption{Trade-Off between performance and acceleration. Model performance and latency reduction consistently vary along with $s$ and $e$.}
    \resizebox{\columnwidth}{!}{
        \begin{tabular}{c|cc|cc}
            \toprule
            \multirow{2}{*}{\begin{tabular}[c]{@{}c@{}} \textbf{$p$} \end{tabular}} & \multirow{2}{*}{\begin{tabular}[c]{@{}c@{}} 
 \textbf{$s$} \end{tabular}} & \multirow{2}{*}{\begin{tabular}[c]{@{}c@{}} \textbf{$e$} \end{tabular}} & \textbf{Downstream Tasks} & \textbf{Latency} \\
             & & & \textbf{Average Score} & \textbf{Reduction} \\
            % \cline{2-5}
            \hline
            \multirow{9}{*}{\begin{tabular}[c]{@{}c@{}} 2 \end{tabular}}  
             & 13 & 30 & 63.2 & 27.0\% \\
            \cline{2-5}
             % \hline
             & 9 & & 60.6 & 32.4\% \\
             & 11 & & 62.0 & 30.3\% \\
             & 15 & & 63.7 & 23.4\% \\
             & 17 & & 63.9 & 21.0\% \\
            \cline{2-5}
             % \hline
             & & 24 & 63.4 & 18.0\% \\
             & & 26 & 63.3 & 20.9\% \\
             & & 28 & 63.3 & 23.9\% \\
             & & 32 & 62.8 & 26.9\% \\
            \bottomrule
        \end{tabular}
    }
    \label{tab:trade_off}
    % \vspace{-0.4cm}
\end{table}

\subsection{Comparison with Tensor Parallelism}
\label{sec:compared_with_tp}

Both tensor parallelism and our method utilize additional GPUs to reduce latency. Therefore, we assess the latency reduction ratio between these two approaches. Given that DeepSpeed inference framework involves optimizations beyond tensor parallelism, it is challenging to make a direct comparison of latency. Therefore, we first measure the latency of both methods using a single GPU under various batch sizes, then evaluate the relative latency reduction ratio with an increasing number of GPUs. The comparison is carried out on the LLaMA-13B model, which is sufficiently large for tensor parallelism to be applied across four GPUs. The maximum batch size of tensor parallelism is 64 due to the limit of available GPU memory.

% TODO: 润色最后一句话

\begin{figure}[t]
    \centering
    \includegraphics[width=\columnwidth]{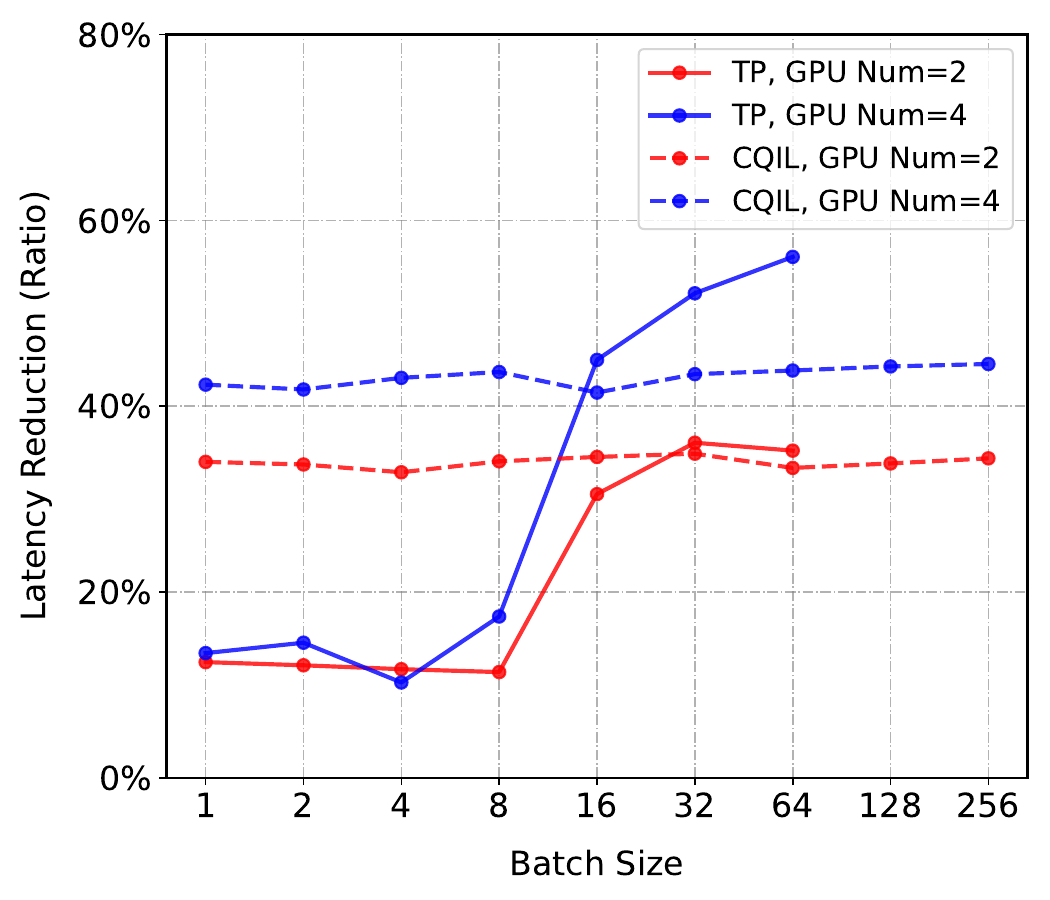}
    \caption{Comparison with Tensor Parallelism. CQIL achieves consistent latency reduction on all batch sizes, benefiting the online inference.}
    \label{fig:comparison_with_tp}
    % \vspace{-0.15cm}
\end{figure}

As shown in Figure \ref{fig:comparison_with_tp}, tensor parallelism does not significantly accelerate the inference for small batch sizes. In contrast, CQIL consistently offers latency reductions for LLaMA-13B, making it more amenable to the scenario of online inference. Moreover, tensor parallelism is complementary to our approach, allowing for the parallel computation of bottom layers, which are sensitive to input substitution. Integrating tensor parallelism with our approach may reduce the latency further. Due to time constraints, we leave the integration of tensor parallelism with CQIL for future work.

\subsection{Combined with Pruning}
\label{sec:combined_with_pruning}

Pruning has been extensively studied as an efficient technique to accelerate model inference. Recently, \citet{sheared-llama} has proposed targeted structured pruning to produce efficient LLMs. Our method is orthogonal to pruning and thus possible to be integrated with it. Table \ref{tab:combined_with_pruning} presents the result of applying CQIL on the Sheared-LLaMA-2.7B model with $p=2, s=13, e=30$, and Sheared-LLaMA-1.3B with $p=2, s=13, e=22$. Results show that our approach achieves further latency reduction for pruned models, proving its effectiveness.

\section{Discussion}
\label{sec:discussion}

Currently, the explanation of the effectiveness of transformer based LLMs can be classified into two views, pipeline \cite{BERTPipeline, BERTStructure, Key-Value-Memories} and ensemble \cite{ResidualEnsembles, Unroll, ViTRobustness}. Based on following observations, our research suggests that LLMs works as a combination of both pipeline and ensemble mechanisms. 

First, our findings reveal that the bottom layers of LLMs display distinct input/output characteristics, rendering them challenging to be parallelized. On the other hand, the middle and top layers demonstrate considerable similarity in their inputs, making it possible for straightforward conversion to parallel execution without sacrificing the model performance. This observation is compatible with previous work \cite{Are_All_Equal}, proving the increased robustness in higher layers within the transformer architecture. 

In other words, our contributions substantiate that LLMs employ a pipeline mechanism at lower levels and an ensemble strategy at higher levels, particularly through the successful parallelization of the middle and upper layers. This work thus provide further evidence to deepen our comprehension on the internal mechanism of LLMs.

\begin{table}[t]
    \centering
    \caption{Combined with Pruning. CQIL achieves further latency reduction while maintaining minimal downstream tasks performance drop.}
    \resizebox{\columnwidth}{!}{
        \begin{tabular}{c|c|c}
            \toprule
            \multirow{2}{*}{\begin{tabular}[c]{@{}c@{}} \textbf{Model} \end{tabular}} & \textbf{Downstream Tasks} & \textbf{Latency} \\
             & \textbf{Average Score} & \textbf{Reduction} \\
            \hline
            \textbf{LLaMA-7B} & 65.0 & 0\% \\
            \hline
            \textbf{Sheared-LLaMA-2.7B} & 58.8 & 55.1\%\\
            \textbf{CQIL-Sheared-LLaMA-2.7B} & 57.4 & 66.0\% \\
            \hline
            \textbf{Sheared-LLaMA-1.3B} & 53.1 & 74.5\% \\
            \textbf{CQIL-Sheared-LLaMA-1.3B} & 52.7 & 78.5\% \\
            \bottomrule
        \end{tabular}
    }
    \label{tab:combined_with_pruning}
    % \vspace{-0.2cm}
\end{table}

\section{Conclusion}

In this paper, we propose an efficient concurrent computation framework on quasi-independent layers that are pervasive in LLMs with the purpose of reducing inference latency while maintaining model performance. To mitigate the potential information loss resulting from input alignment and improve the performance, we develop the bypassing technique, transmitting attention outputs among input-aligned layers. Our experimental results justify the effectiveness of CQIL method. In the future, we will apply CQIL on even larger models and implement the integration of tensor parallelism with our approach to achieve further latency reductions.

 \section*{Limitations}

This work has three primary limitations. First, CQIL requires additional GPUs to reduce the inference latency, limiting its application in the environment equipped with only a single GPU. Second, not every layer in LLMs could be computed concurrently, especially for layers at the bottom and top. These layers are computed on a single GPU, leaving remaining GPUs ignored. To efficiently utilize all GPUs, we suggest to apply tensor parallelism for these layers. Due to time constraints, the integration of tensor parallelism and our approach is leaved for future work. Finally, the computing FLOPS remain unchanged. Therefore, from an energy consumption perspective, CQIL does not result in electricity savings. 

% \input{tex/acknowledge}

% Bibliography entries for the entire Anthology, followed by custom entries
%\bibliography{anthology,custom}
% Custom bibliography entries only
\bibliography{acl}

\appendix

\section{Attention-FFN Parallelism}

Given that both the attention and feedforward modules use pre-layer normalization, aligning their inputs is feasible. However, the computation times for these two modules usually do not match, leading to one GPU being underutilized if they are processed concurrently on separate GPUs. Therefore, attention-ffn parallelism is not employed in our study. Nevertheless, parallelism between the attention and feedforward modules holds potential for acceleration, especially when two computational resources are available but their speeds differ. For example, \citet{PowerInfer} proposes to utilize both CPU and GPU for LLMs inference on personal computers. The formulation of attention-ffn parallelism is shown in Eq.\ref{eq:attn_ffn_parallelism_input_formulation}.

\begin{align}
    x_{l+1} = x_{l} &+ \text{ATTN}_{l}(x_{l}) + \text{FFN}_{l}(x_{l})
    \label{eq:attn_ffn_parallelism_input_formulation}
\end{align}

Our empirical experiments show that, the attention and feedforward modules can be computed concurrently in most layers. Specifically, we parallelize attention and feedforward modules for layers ranging from $13$ to $30$ in the LLaMA-7B model and achieve an average downstream task score of 63.7 (compared to the original model's score of 65.0). Such parallelism may enhance the efficiency of existing CPU-GPU collaborative inference frameworks, paving the way for further acceleration.

\section{Pre-training Parallel Architecture}

\begin{table}[t]
    \centering
    \caption{Pre-training Parallel Architecture.}
    \label{tab:architecture_of_pl}
    \resizebox{\columnwidth}{!}{
        \begin{tabular}{c|ccc}
            \toprule
            \textbf{Model} & \textbf{LAMBADA} & \textbf{PTB} & \textbf{WikiText103} \\
            \hline
            GPT2 & 70.8 & 139.4 & 56.2 \\
            CQIL-FT-GPT2 & 84.6 & 156.2 & 73.5 \\
            CQIL-Pretrained-GPT2 & 84.5 & 189.1 & 81.4\\
            \bottomrule
        \end{tabular}
    }
\end{table}

Our work focuses on converting pre-trained LLMs into a model with parallel layers. An alternative strategy is to pre-train an architecture with parallel layers from scratch, potentially yielding comparable results and latency improvements. We explored this direction through experiments. We first pre-trained a 12 layers GPT-2\cite{gpt2} model with 100K steps on Wikipedia and BookCorpus\cite{BookCorpus} datasets. Then the model was converted to the parallel architecture with $p=2, s=3, e=12$ and fine-tuned with 5K steps. Meanwhile, we pre-trained the parallel architecture with 105K steps from scratch. We evaluate zero-shot perplexities on LAMBADA, PTB and WikiText103 as the performance of downstream tasks. Results in Table \ref{tab:architecture_of_pl} indicate that training a parallel architecture from scratch does not achieve the downstream task performance attained by fine-tuning the transferred model. This performance gap may arise from the enhanced fitting capabilities inherent to deeper models during pre-training. The finding underscores the critical role of transitioning from pre-trained LLMs to the parallel architecture.

\section{Details of Fine-tuning}
\label{sec:finetune_details}

\begin{table}[h]
    \centering
    \caption{Fine-tuning Details.}
    \label{tab:fine_tuning_details}
    \resizebox{\columnwidth}{!}{
        \begin{tabular}{c|ccc|cc}
            \toprule
             \multirow{2}{*}{\begin{tabular}[c]{@{}c@{}} \textbf{Model} \end{tabular}} & \multirow{2}{*}{\begin{tabular}[c]{@{}c@{}} \textbf{$p$} \end{tabular}} & \multirow{2}{*}{\begin{tabular}[c]{@{}c@{}} \textbf{$s$} \end{tabular}} & \multirow{2}{*}{\begin{tabular}[c]{@{}c@{}} \textbf{$e$} \end{tabular}} & \textbf{Learning} & \textbf{LoRA} \\
             &  &  &  & \textbf{Rate} & \textbf{Rank} \\
            \hline
            \textbf{Sheared-LLaMA-1.3B} & 2 & 13 & 22 & 1e-5 & 2 \\
            \textbf{Sheared-LLaMA-2.7B} & 2 & 13 & 30 & 1e-5 & 2 \\
            \hline
            \multirow{2}{*}{\begin{tabular}[c]{@{}c@{}} \textbf{LLaMA-7B} \end{tabular}}
             & 2 & 13 & 30 & 1e-5 & 2 \\
             & 4 & 15 & 30 & 1e-5 & 32 \\
            \hline
            \multirow{2}{*}{\begin{tabular}[c]{@{}c@{}} \textbf{LLaMA-13B} \end{tabular}}
             & 2 & 11 & 38 & 1e-4 & 64 \\
             & 4 & 15 & 38 & 1e-4 & 64 \\
            \hline
            \multirow{2}{*}{\begin{tabular}[c]{@{}c@{}} \textbf{LLaMA-33B} \end{tabular}}
             & 2 & 11 & 58 & 1e-5 & 64 \\
             & 4 & 19 & 58 & 1e-5 & 64 \\
            \bottomrule
        \end{tabular}
    }
\end{table}

We fine-tune all models with context length of 2,048, weight decay of 0.1, batch size of 32, dropout probability of 0, warmup steps of 0 and steps of 8000. The learning rate is constant and Adam optimizer is used. For simplicity, lora alpha is always equal to lora rank. We fine-tune all linear matrices without biases in attention and feedforward modules. Mixed precision of bf16 and DeepSpeed\cite{DeepSpeed} framework are used for fine-tuning. More details could be found in Table \ref{tab:fine_tuning_details}.

\section{The Potential Performance Degradation of Pruning}
\label{sec:the_potential_harm_of_pruning}

\begin{figure}[h]
    \setlength{\belowcaptionskip}{-0.35cm}
    \centering
    \includegraphics[width=\columnwidth]{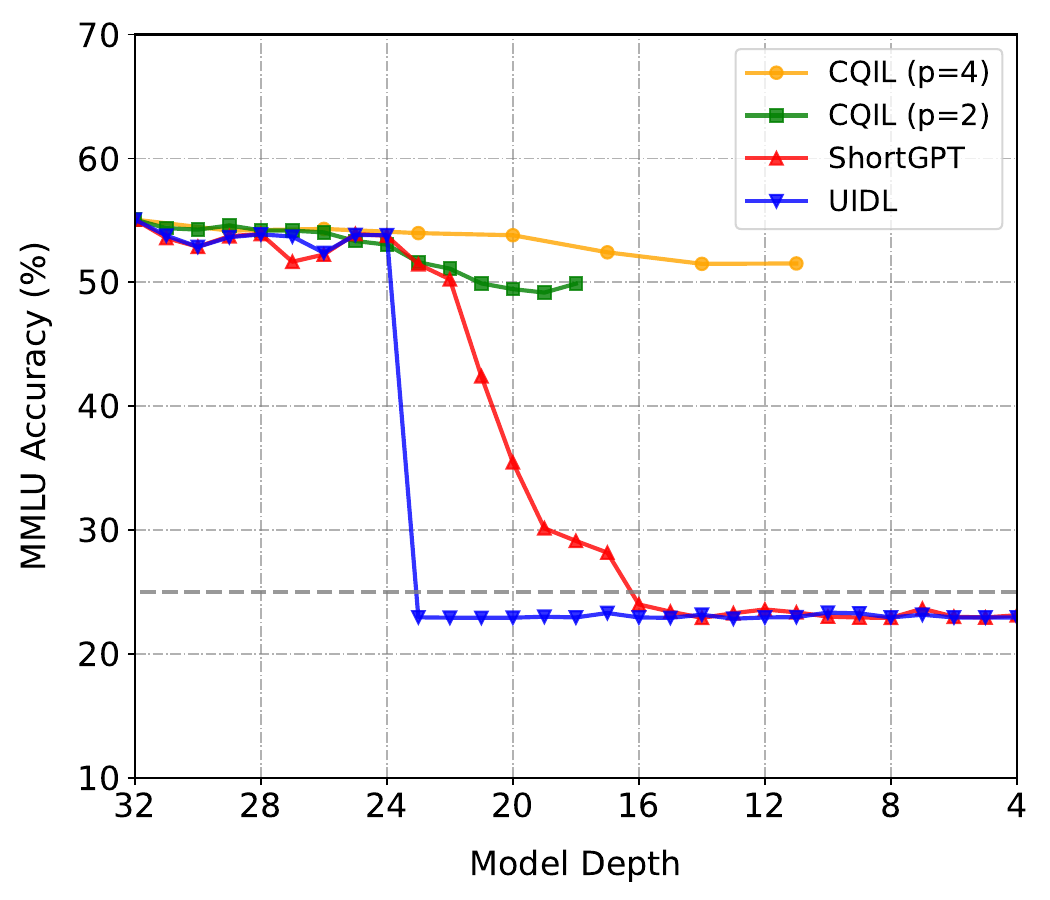}
    \caption{MMLU (zero-shot) comparison with similarity based pruning methods. Dashed gray line represents the random guessing score.}
    \label{fig:mmlu_comparison_with_pruning}
    % \vspace{-0.15cm}
\end{figure}

During the review phase of our manuscript, two similarity based pruning methods\cite{ShortGPT,Unreasonable} were posted. The findings reported in these studies indicate that the impact on performance is negligible when pruning up to half of the layers in LLMs. Considering the shared motivation but distinct approaches of these methods, we conduct the comparative experiments to demonstrate the potential problems of pruning.

\begin{figure}[t]
    \centering
    \includegraphics[width=\columnwidth]{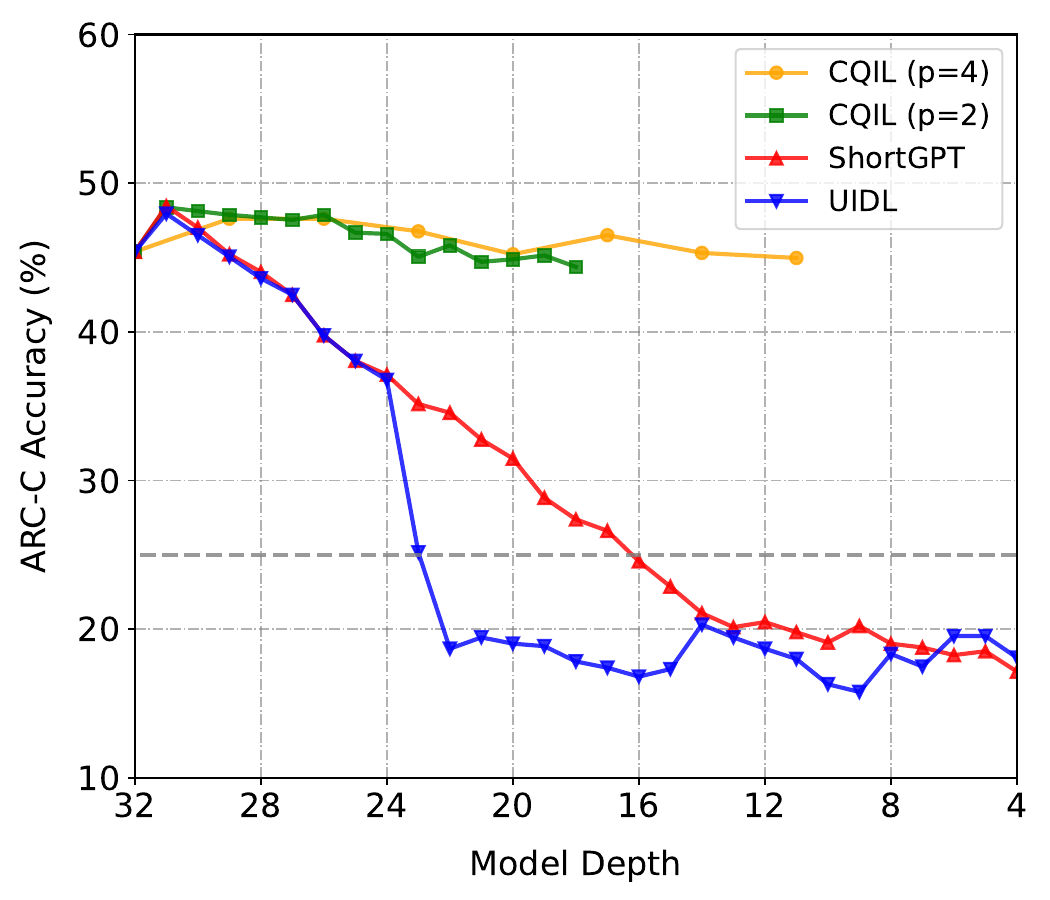}
    \caption{ARC-Challenge (zero-shot) comparison with similarity based pruning methods. Dashed gray line represents the random guessing score.}
    \label{fig:arc-c_comparison_with_pruning}
    % \vspace{-0.15cm}
\end{figure}

\begin{figure}[t]
    \centering
    \includegraphics[width=\columnwidth]{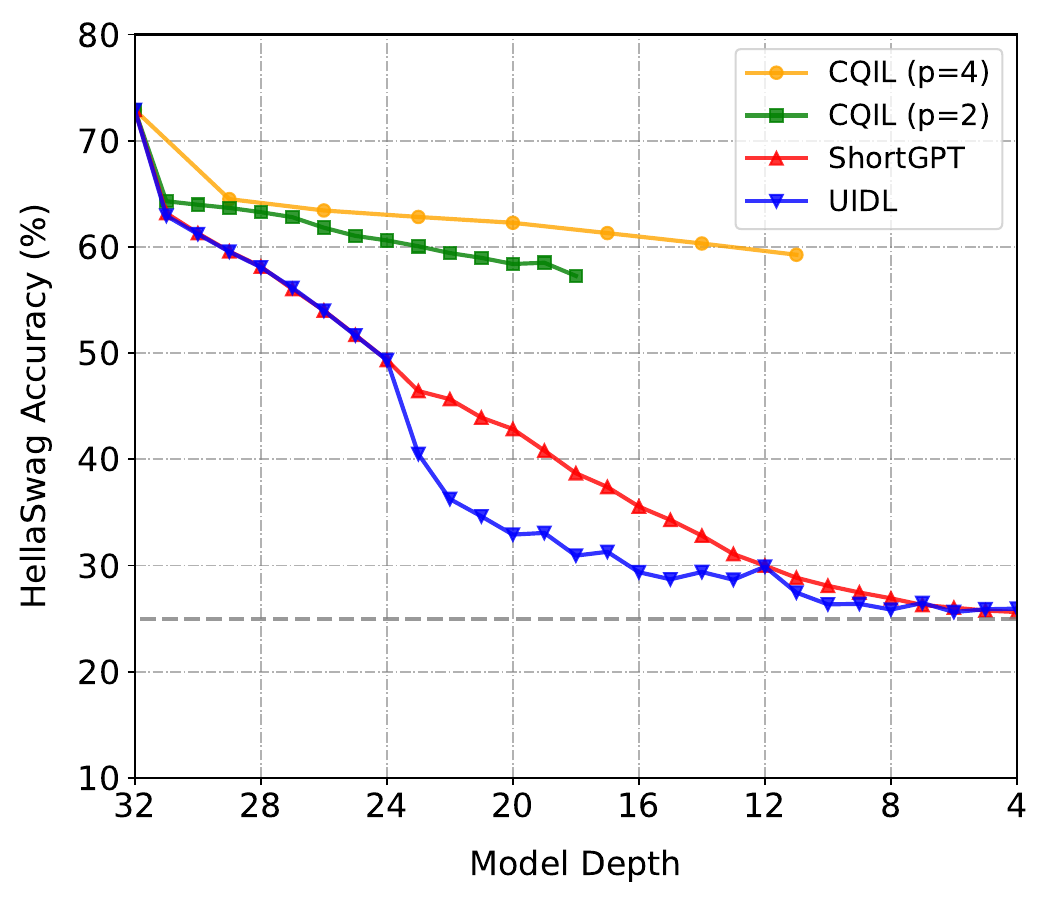}
    \caption{HellaSwag (zero-shot) comparison with similarity based pruning methods. Dashed gray line represents the random guessing score.}
    \label{fig:hellaswag_comparison_with_pruning}
    % \vspace{-0.15cm}
\end{figure}

\begin{table*}[t]
    \centering
    \caption{Downstream tasks performance of models without fine-tuning.}
    \resizebox{\textwidth}{!}{
        \begin{tabular}{cccccccccc}
            \toprule
            \multirow{2}{*}{\begin{tabular}[c]{@{}c@{}} \textbf{Model} \end{tabular}} & \multicolumn{3}{c}{\textbf{Partition Strategy}} & \multicolumn{6}{c}{\textbf{Commonsense\&Reading Comprehension}} \\
            % \cline{3-8}
             & $p$ & $s$ & $e$ & \textbf{SciQ} & \textbf{PIQA} & \textbf{WinoGrande} & \textbf{ARC-E} & \textbf{ARC-C} & \textbf{HellaSwag} \\
            \hline
            \multirow{3}{*}{\begin{tabular}[c]{@{}c@{}} \textbf{LLaMA-7B} \end{tabular}} 
             & 1 & 1 & 32 & 93.0 & 79.2 & 70.0 & 72.9 & 44.8 & 76.2 \\
             & 2 & 13 & 30 & 89.8 & 77.3 & 67.6 & 68.4 & 41.3 & 54.9 \\
             & 4 & 15 & 30 & 87.4 & 75.7 & 65.7 & 63.1 & 41.0 & 70.5 \\
            \hline
            \multirow{3}{*}{\begin{tabular}[c]{@{}c@{}} \textbf{LLaMA-13B} \end{tabular}} 
             & 1 & 1 & 40 & 91.3 & 80.1 & 72.8 & 74.8 & 47.6 & 79.1 \\
             & 2 & 11 & 38 & 87.9 & 78.5 & 69.5 & 70.2 & 44.5 & 76.1 \\
             & 4 & 15 & 38 & 86.7 & 77.9 & 69.0 & 64.5 & 41.6 & 73.8 \\
             \hline
            \multirow{3}{*}{\begin{tabular}[c]{@{}c@{}} \textbf{LLaMA-33B} \end{tabular}} 
             & 1 & 1 & 60 & 94.6 & 82.3 & 76.0 & 79.0 & 52.1 & 82.6 \\
             & 2 & 11 & 58 & 90.8 & 78.5 & 70.7 & 68.9 & 46.2 & 77.2 \\
             & 4 & 19 & 58 & 90.8 & 79.3 & 72.6 & 70.7 & 47.3 & 78.7 \\
             \midrule
            \multirow{2}{*}{\begin{tabular}[c]{@{}c@{}} \textbf{Model} \end{tabular}} & \multicolumn{3}{c}{\textbf{Partition Strategy}} & \multicolumn{2}{c}{\textbf{Continued}} & \textbf{LM} & \textbf{World Knowledge} & \multicolumn{2}{c}{\textbf{Downstream Tasks}} \\
              & $p$ & $s$ & $e$ & \textbf{LogiQA} & \textbf{BoolQ} & \textbf{LAMBADA} & \textbf{MMLU (5)} & \multicolumn{2}{c}{\textbf{Average Score}} \\
            \hline
            \multirow{3}{*}{\begin{tabular}[c]{@{}c@{}} \textbf{LLaMA-7B} \end{tabular}}
             & 1 & 1 & 32 & 30.0 & 75.1 & 73.5 & 35.1 & \multicolumn{2}{c}{65.0} \\
             & 2 & 13 & 30 & 29.8 & 73.4 & 68.5 & 31.4 &  \multicolumn{2}{c}{60.2}\\
             & 4 & 15 & 30 & 27.5 & 44.7 & 56.7 & 29.8 & \multicolumn{2}{c}{56.2} \\
            \hline
            \multirow{3}{*}{\begin{tabular}[c]{@{}c@{}} \textbf{LLaMA-13B} \end{tabular}} 
             & 1 & 1 & 40 & 32.0 & 77.9 & 76.2 & 46.7 & \multicolumn{2}{c}{67.8} \\
             & 2 & 11 & 38 & 31.6 & 73.2 & 69.4 & 39.3 & \multicolumn{2}{c}{64.0} \\
             & 4 & 15 & 38 & 28.6 & 63.7 & 60.1 & 44.6 & \multicolumn{2}{c}{61.1} \\
             \hline
            \multirow{3}{*}{\begin{tabular}[c]{@{}c@{}} \textbf{LLaMA-33B} \end{tabular}} 
             & 1 & 1 & 60 & 31.8 & 82.6 & 77.6 & 58.2 & \multicolumn{2}{c}{71.7} \\
             & 2 & 11 & 58 & 23.8 & 77.1 & 71.3 & 47.0 & \multicolumn{2}{c}{65.2} \\
             & 4 & 19 & 58 & 25.4 & 80.4 & 61.3 & 45.7 & \multicolumn{2}{c}{65.2} \\
            \bottomrule
        \end{tabular}
    }
    \label{tab:wo_finetune_experiment_results}
\end{table*}

\begin{table*}[t]
    \centering
    \caption{Numerical values of input substitution experiments (Layer 1-16).}
    \resizebox{\textwidth}{!}{
        \begin{tabular}{c|c|c|c|c|c|c|c|c|c|c|c|c|c|c|c|c}
            \toprule
            Layer & 1 & 2 & 3 & 4 & 5 & 6 & 7 & 8 & 9 & 10 & 11 & 12 & 13 & 14 & 15 & 16 \\
            \hline
            k=1 &  & 868.2 & 6.4 & 14.1 & 6.1 & 6.0 & 6.0 & 6.0 & 6.0 & 6.0 & 6.0 & 6.0 & 6.0 & 6.0 & 6.0 & 6.0 \\
            k=3 &  &  &  & 16.3 & 13.7 & 11.2 & 6.2 & 6.2 & 6.1 & 6.1 & 6.1 & 6.1 & 6.1 & 6.1 & 6.1 & 6.1 \\
            k=7 &  &  &  &  &  &  &  & 8.8 & 8.9 & 8.3 & 6.3 & 6.3 & 6.3 & 6.2 & 6.2 & 6.2 \\
            \bottomrule
        \end{tabular}
    }
    \label{tab:numerical_misin_1}
\end{table*}

\begin{table*}[t!]
    \centering
    \caption{Numerical values of input substitution experiments (Layer 17-32).}
    \resizebox{\textwidth}{!}{
        \begin{tabular}{c|c|c|c|c|c|c|c|c|c|c|c|c|c|c|c|c}
            \toprule
            Layer & 17 & 18 & 19 & 20 & 21 & 22 & 23 & 24 & 25 & 26 & 27 & 28 & 29 & 30 & 31 & 32 \\
            \hline
            k=1 & 6.0 & 6.0 & 6.0 & 6.0 & 6.0 & 5.9 & 5.9 & 5.9 & 5.9 & 5.9 & 5.9 & 5.9 & 6.0 & 6.0 & 6.0 & 7.5 \\
            k=3 & 6.1 & 6.1 & 6.0 & 6.0 & 6.0 & 6.0 & 6.0 & 6.0 & 6.0 & 6.0 & 6.0 & 6.0 & 6.1 & 6.1 & 6.4 & 8.7 \\
            k=7 & 6.4 & 6.3 & 6.2 & 6.2 & 6.2 & 6.2 & 6.2 & 6.2 & 6.1 & 6.2 & 6.2 & 6.2 & 6.7 & 6.5 & 6.9 & 10.6 \\
            \bottomrule
        \end{tabular}
    }
    \label{tab:numerical_misin_2}
\end{table*}

\begin{figure*}[t]
    \centering
    \includegraphics[width=\textwidth]{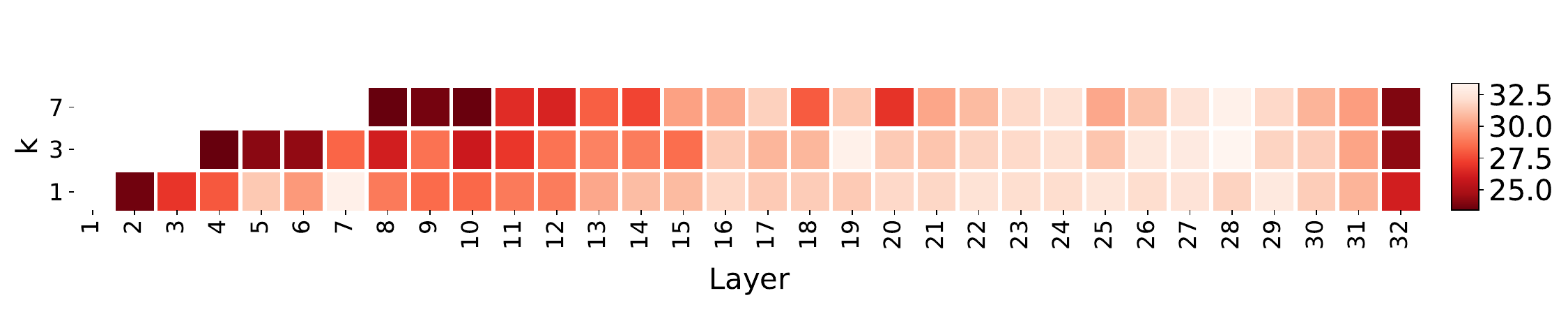}
    \caption{MMLU Sensitivity of Input Substitution. We individually replace the input of layer $l$ with that of the layer $l-k$ and evaluate the MMLU score with zero-shot setting. A darker block indicates a lower MMLU score and diminished performance. Note that the zero-shot MMLU score of the original model is 33.4.}
    \label{fig:misin_mmlu}
\end{figure*}

Specifically, we applied the two layer pruning methods for Qwen-7B\cite{Qwen}. Subsequently, the pruned models were fine-tuned using LoRA with rank of $64$, constant learning rate of 1e-4, batch size of $32$ context length of 2,048, and steps of $8000$. For CQIL, we set $e=30, d=1$ and progressively decrease the $s$ to obtain models with different depth. The CQIL applied models are fine-tuned under identical settings. For ease of reference, we denote these two approaches as ShortGPT\cite{ShortGPT} and UIDL\cite{Unreasonable}, respectively. To assess the performance of downstream tasks, the fine-tuned models are evaluated on MMLU, ARC-Challenge and HellaSwag datasets in a zero-shot setting.

The result, as illustrated in Figure \ref{fig:mmlu_comparison_with_pruning}, demonstrates that the MMLU score has minimal degradation with the pruning of fewer than 9 layers, suggesting the potential redundancy of LLMs. However, as shown in Figures \ref{fig:arc-c_comparison_with_pruning} and \ref{fig:hellaswag_comparison_with_pruning}, the performance of pruned models consistently decline on other benchmark datasets. These observations suggest that while the MMLU dataset is substantial, it may not fully capture the comprehensive capabilities or knowledge inherent to LLMs. Consequently, we argue for the necessity of retaining LLM layers to maintain their true functional potential.

\section{Downstream Tasks Performance of CQIL-Models without Fine-Tuning}

Without additional training, models with CQIL still preserve a close level of performance to the original. Results without fine-tuning are shown in Table \ref{tab:wo_finetune_experiment_results}.

\section{Additional Details of Input Substitution Experiments}

To clarify the details of Figure \ref{fig:misin}, we give the numerical values of each substitution experiment in Table \ref{tab:numerical_misin_1} and \ref{tab:numerical_misin_2}. Moreover, we conducted the sensitivity experiment on LLaMA-7B with MMLU under zero-shot setting. The results in Table \ref{fig:misin_mmlu} are similar to Figure \ref{fig:misin}, revealing that the majority of middle layers are relatively insensitive. 

\section{Additional Bypassing Experiments}

\begin{table}[h]
    \centering
    \caption{Effect of bypassing on LLaMA-7B with $p=2$.}
    \resizebox{\columnwidth}{!}{
        \begin{tabular}{c|c|c}
            \toprule
            \textbf{Bypassing} & \textbf{Downstream Tasks} & \textbf{Latency} \\
            \textbf{Distance ($d$)} & \textbf{Average Score} & \textbf{Reduction} \\
            \hline
             0 & 62.6 & 27.6\% \\
             1 & 63.2 & 27.0\% \\
            \bottomrule
        \end{tabular}
    }
    \label{tab:add_effect_of_bypassing}
    % \vspace{-0.4cm}
\end{table}

We conducted additional experiments for LLaMA-7B with $p=2$. The results in Table \ref{tab:add_effect_of_bypassing} confirm that bypassing technique improves the performance on all LLaMA-7Bs with different $p$ values.

\end{document}